%% file: arxiv_main.tex
\documentclass[a4paper]{svproc}

\input{preamble.tex}
\usepackage{url}

\begin{document}

\definecolor{darkred}{RGB}{200,0,0}
\begin{textblock*}{150mm}(30mm,20mm)
{\color{darkred}\begin{center}\textbf{Accepted in The 19th International Symposium on Experimental Robotics – ISER 2025\\6-10 July 2025, Santa Fe, New Mexico, USA}\end{center}}
\end{textblock*}

\mainmatter              
%
\title{\texorpdfstring{Learning to Optimize Package Picking for\\ Large-Scale, Real-World Robot Induction}{Learning to Optimize Package Picking for Large-Scale, Real-World Robot Induction}}
\titlerunning{Learning to Optimize Package Picking}  
%
\author{Shuai Li\inst{1} \and Azarakhsh Keipour\inst{1}
\and Sicong Zhao\inst{1}
\and Srinath Rajagopalan\inst{1}
\and \\ Charles Swan\inst{1}
\and Kostas Bekris\inst{1,2}}
\authorrunning{Shuai Li et al.} 
%
\tocauthor{Shuai Li, Azarakhsh Keipour, Sicong Zhao, Srinath Rajagopalan, Charles Swan, Kostas E. Bekris}
\institute{Amazon Robotics, Seattle WA 98109, USA,\\
\email{shuailirpi@gmail.com, keipour@gmail.com, \\ \{sczhao,rajsrina,cswan\}@amazon.com},\\
\and
Rutgers University, Piscataway, New Jersey 08854, USA,
\\
\email{kostas.bekris@cs.rutgers.edu}}

\maketitle              

\vspace{-.20in}
\begin{abstract}
\input{sections/abstract}
\vspace{-0.05in}
\end{abstract}

\vspace{-0.05in}
\input{sections/01_intro}

\input{sections/02_approach}
\input{sections/03_experiments}
\input{sections/04_insights}

\bibliographystyle{spmpsci.bst}
\bibliography{references,che_references}
\end{document}

%% file: preamble.tex
\usepackage{amssymb,amsmath,amsfonts,mathrsfs}
\usepackage{graphicx}
\usepackage[table,xcdraw,dvipsnames]{xcolor}
\usepackage{tikz}

\usepackage[labelformat=simple]{subcaption}

\usepackage{colortbl}
\usepackage{booktabs}
\usepackage{array}
\usepackage{tabularx}
\usepackage{wrapfig}
\usepackage{makecell}

\usepackage[normalem]{ulem}

\usepackage[absolute]{textpos}

\usepackage[flushleft]{threeparttable}
\usepackage{multirow}

\newcolumntype{L}[1]{>{\raggedright\let\newline\\\arraybackslash\hspace{0pt}}m{#1}}

\usepackage{stackengine}

\usepackage{textcomp} 
\usepackage{stfloats}
\usepackage[hyphens]{url}
\usepackage{verbatim}
\usepackage{graphicx}
\usepackage[dvipsnames]{xcolor}
\usepackage[numbers]{natbib}
\usepackage{times}
\usepackage{multicol}
\PassOptionsToPackage{hyphens}{url}
\usepackage{hyperref}

\usepackage{enumitem}
\usepackage{tabularx}

\usepackage{booktabs}       

\usepackage{arydshln}
\makeatletter
\def\adl@drawiv#1#2#3{%
        \hskip.5\tabcolsep
        \xleaders#3{#2.5\@tempdimb #1{1}#2.5\@tempdimb}%
                #2\z@ plus1fil minus1fil\relax
        \hskip.5\tabcolsep}
\newcommand{\cdashlinelr}[1]{%
  \noalign{\vskip\aboverulesep
           \global\let\@dashdrawstore\adl@draw
           \global\let\adl@draw\adl@drawiv}
  \cdashline{#1}
  \noalign{\global\let\adl@draw\@dashdrawstore
           \vskip\belowrulesep}}
\makeatother

\usepackage{algorithm}
\usepackage[noend]{algpseudocode}
\definecolor{commentcolor}{gray}{0.5}
\algnewcommand{\LineComment}[1]{\State \textcolor{commentcolor}{\(\triangleright\) #1}}
\algnewcommand{\var}[1]{\textit{#1}}
\algnewcommand{\func}[1]{\textsc{#1}}

\DeclareMathOperator*{\argmax}{arg\,max}

{\end{list}}

%% file: sections/abstract.tex
Warehouse automation plays a pivotal role in enhancing operational efficiency, minimizing costs, and improving resilience to workforce variability. While prior research has demonstrated the potential of machine learning (ML) models to increase picking success rates in large-scale robotic fleets by prioritizing high-probability picks and packages, these efforts primarily focused on predicting success probabilities for picks sampled using heuristic methods. Limited attention has been given, however, to leveraging data-driven approaches to directly optimize sampled picks for better performance at scale. In this study, we propose an ML-based framework that predicts transform adjustments as well as improving the selection of suction cups for multi-suction end effectors for sampled picks to enhance their success probabilities. The framework was integrated and evaluated in test workcells that resemble the operations of Amazon Robotics' Robot Induction (Robin) fleet, which is used for package manipulation. Evaluated on over 2 million picks, the proposed method achieves a 20\% reduction in pick failure rates compared to a heuristic-based pick sampling baseline, demonstrating its effectiveness in large-scale warehouse automation scenarios.

%% file: sections/01_intro.tex
\vspace{-.3in}
\section{Motivation, Problem Statement and Related Work} 
\vspace{-.1in}
\label{sec:intro}

\begin{wrapfigure}{r}{0.55\textwidth}
    \centering
    \vspace{-.3in}
    \includegraphics[width=0.2\textwidth, height=3.8cm]{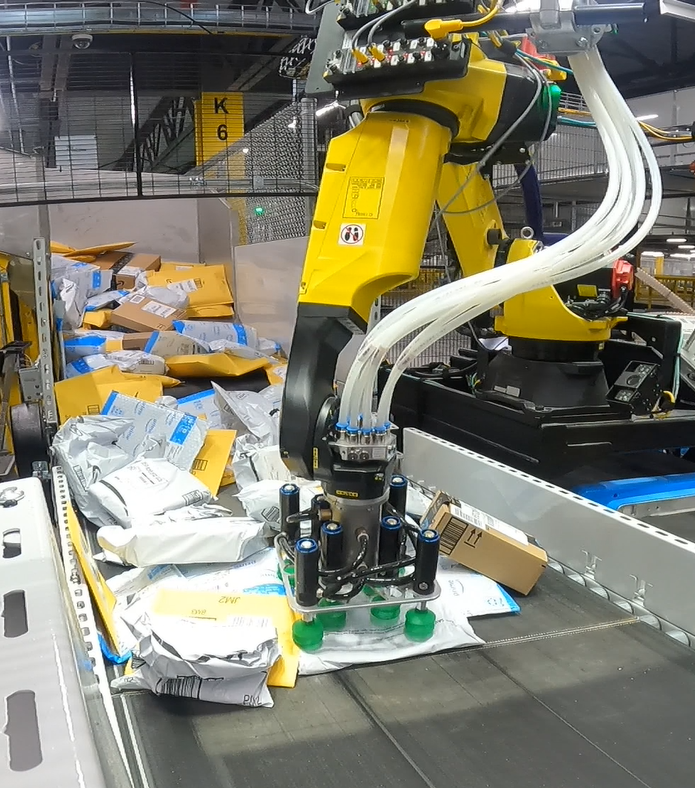}%
    ~
    \includegraphics[width=0.34\textwidth, height=3.8cm]{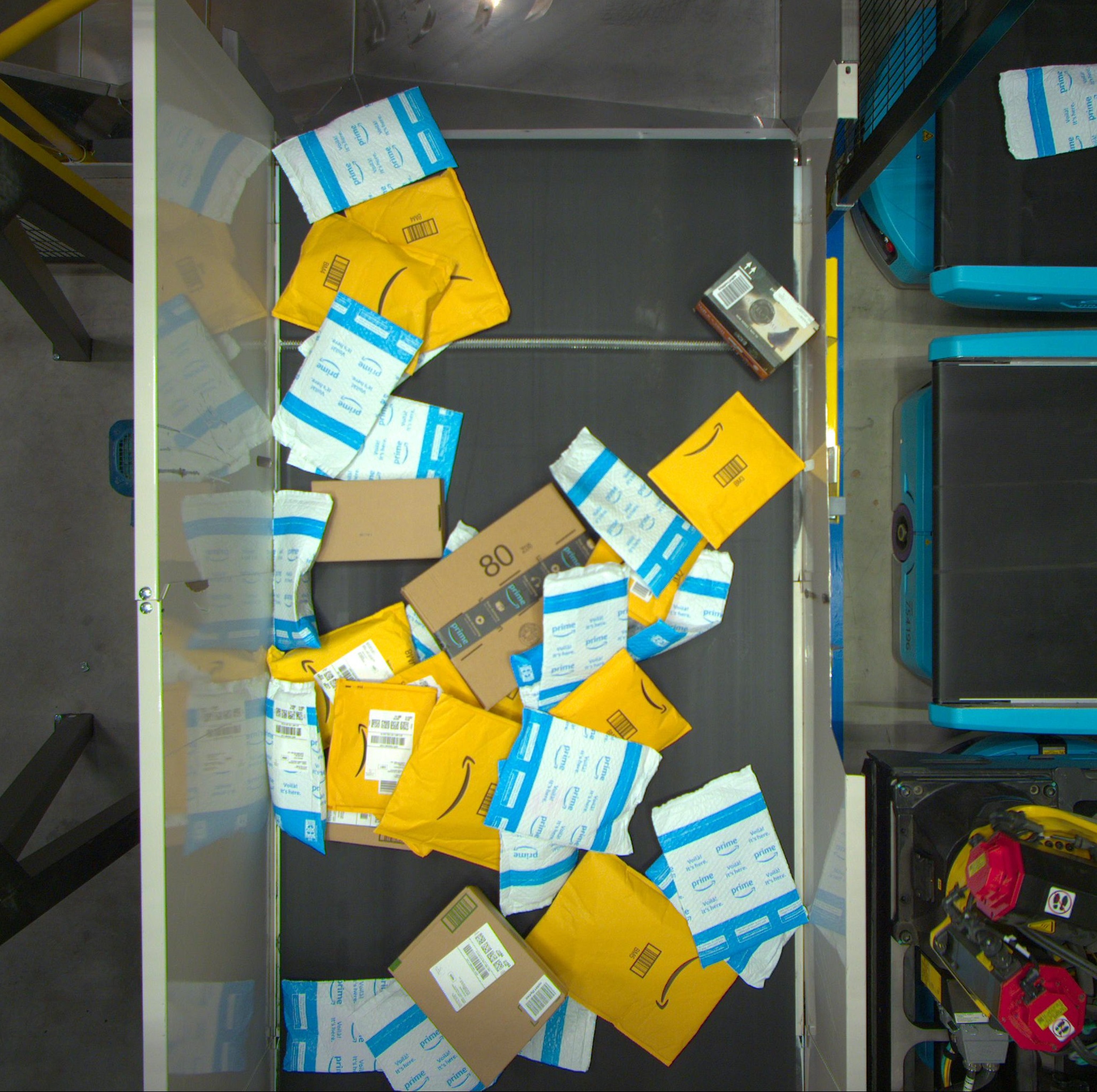}%
    \vspace{-.1in}
    \caption{\small An example of a Robot induction (Robin) workcell. Robin performs automated package singulation by picking packages from unstructured piles on conveyors and transferring them to mobile robots. }
    \vspace{-.3in}
    \label{fig:robin-workcell}
\end{wrapfigure}


Robot induction systems, such as the one shown in Fig.~\ref{fig:robin-workcell}, which automate singulation and sorting in logistics, face multiple challenges. They include high object diversity, deformable material, complex pile configurations, and significant occlusions. Despite significant progress in computer vision and machine learning, identifying effective picks remains a critical bottleneck for effective, real-world industrial operations~\cite{robin}. In particular, improvements in pick success rates can translate to significant operational gains in large-scale industrial settings. 

{\bf Prior Work in Robot Picking} Traditional methods for robot grasping \cite{graspit} involve geometric reasoning and planning typically for multi-fingered hands. They often calculate the poses and forces for robotic contacts to satisfy mechanical constraints. The requirement, however, for accurate geometric and physical object models often limits their applicability in unstructured setups involving open-sets of objects ~\cite{grasp-survey}. 

This has motivated the development of learning-based picking approaches \cite{dl_robot_grasping}. Such data-driven methods brought the promise of higher effectiveness in challenging, cluttered setups with unknown objects. Morrison et al.~\cite{gg-cnn} estimated grasp quality from depth images, while Jiang et al.~\cite{geometric} explored geometric, feature-based approaches. Most ML methods are  focused on parallel, pinch grippers \cite{fang2023anygrasp, fang2020graspnet, yuan2023m2t2} or are dealing with simple setups, such as single-object picking. The Amazon Picking Challenge \cite{apc_paper}, however, demonstrated that suction-based grippers can be very effective in real-world picking setups by simplifying pick reasoning. Suction-based grippers have been deployed in real-world production environments and allow fast and robust picking of items that can potentially be heavy. 

Some of the data-driven solutions for robot picking have been extended to address suction-based grippers, such as the Dex-Net solutions, which have addressed open-ended grasp quality prediction from point clouds \cite{mahler2018dex, suctionnet}. In Dex-Net, a pick candidate is evaluated using an expert-designed evaluation system. These promising approaches, however, may not fully address the complexity and tight operational constraints of industrial settings. Researchers have explored interactive strategies for scenarios where picks are not immediately available~\cite{affordance}, but these approaches tend to be too time-consuming for practical applications.

The current work assumes access to a model that has learned pick success prediction given success labels of past production data ~\cite{Li:2023:rss:pickranking, Li:2023:iros-workshop:pick-strategies}. This prior work by Li et al. has explored shallow, explainable models trained on large-scale, real-world data. More recent work in learning pick success models for multi-suction picks \cite{Che:2025:rss:deeppick} has shown that with proper pretraining and finetuning, a deep model with a multimodal visual encoder can in fact outperform the shallow model as well as alternative deep architectures in terms of picking items from an open-set. 


{\bf Contribution} This work presents a novel approach that focuses on optimizing candidate picks given access to a model of pick success probability. The input picks can be generated either via existing heuristic or learned solutions. The approach refines the parameters of the input picks by using a machine learning model that is trained on small-scale data gathered from physical robots and utilizing the learned model of pick success. The proposed optimization model outputs higher-quality picks that improve success rates. We demonstrate effectiveness via comprehensive evaluation over two million robotic inductions performed on test workcells that resemble Amazon warehouse operations for robot induction.

Overall, the key contributions include: (1) \textbf{Pick Optimization Framework:} A novel method for adjusting picks to enhance their success probability given access to few physical data and a prior model for evaluating pick success; (2) \textbf{Demonstrable Improvement:} Enhanced performance over ranking-based approaches based on the same underlying model for evaluating pick success that rank multiple samples, validated through extensive A/B testing involving 2 million picks. 



%% file: sections/02_approach.tex
\section{Technical Approach} 
\label{sec:approach}




Consider the picking task illustrated in Fig.~\ref{fig:robin-workcell}. The task is initiated when a picking scene $s_t$ of cluttered packages of different types arrives at a reachable area via a conveyor belt. The conveyor belt and the scene remain static throughout the picking process. The picking is performed by an induction manipulation robot consisting of a multiple-DoF arm with a end-of-arm tool (EoAT) that can activate multiple suction cups.

Each pick action $a$ is defined as a set of variables determining: a 3-D point in space (i.e., the desired pick point where the EoAT makes contact with an item’s surface), the desired 3-D orientation of the EoAT at the pick point, and a set of desired active suction cups on the EoAT. 

A feature vector $\phi(s_t, a) \in \mathbb{R}^d$ encodes both pick action $a$ parameters and scene $s_t$ information, such as the geometric relationship between the end-of-the-arm tool (EoAT) and nearby surfaces. A function $F : \mathbb{R}^d \rightarrow [0,1]$ maps these features $\phi(s_t, a)$ to pick success probability. 

{\bf Access to Learned Model of Pick Success:} Previous research demonstrated machine learning models that approximate the pick success probability function $F$ using historical data from executed picks in industrial settings~\cite{Li:2023:rss:pickranking}. This work assumes access to such learned models of $F$.





{\bf Optimization Objective:} For an initial candidate pick $a$, its local neighborhood $\mathcal{N}(a)$, and the scene $s_t$, our approach incrementally refines the pick parameters to maximize the predicted success probability, which can be expressed as finding $a^*$ as follows:

\begin{equation*}
    a^* = \argmax_{a' \in \mathcal{N}(a)} F(\phi(s_t, a'))
\end{equation*}

\begin{wrapfigure}{r}{0.45\textwidth}
    \vspace{-.3in}
    \centering
    \includegraphics[width=0.99\linewidth]{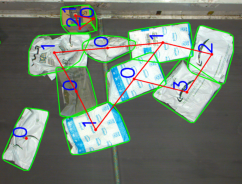}
    \vspace{-.1in}
    \caption{Example of an adjacency graph for a cluster of items.}
    \label{fig:adj-graph}
    \vspace{-.2in}
\end{wrapfigure}

\noindent {\bf Features:} The features $\phi(s_t, a)$ can be shared between the model of pick success $F(\phi(s_t, a'))$ and the proposed optimization model that refines pick parameters. We compute a set of features for each induct using scene metadata as well as RGB, and depth images. Specifically, the camera data is processed to generate package segments and tag each segment with an associated package type label. Additional statistics are computed for each segment using depth information (e.g., surface normals and the quality of plane fitting). The following features have been shown to be good predictors of pick quality:

\begin{itemize}
\item Package height: This feature correlates with the
package’s momentum and can impact the shear
force at the suction cups and the pick’s stability.
\item Quality of plane fitting: A plane is fit on each segment;
a better plane fit correlates with a better seal between the suction cups and the package.
\item Number of activated suction cups: More active suction
cups means a more stable pick, reducing the failure
probability.
\item Alignment quality between the suction cups and the
package surface: This feature is computed as the offsets
between the package surface normal vector and the normal
vector of the suction cups. A better alignment indicates a better seal between the suction cups and the package surface.
\end{itemize}

\begin{figure}[!htb]
    \centering
    \vspace{-.2in}
    \begin{subfigure}{0.40\linewidth}
        \stackunder[1pt]{\includegraphics[width=0.499\textwidth, height=2.2cm]{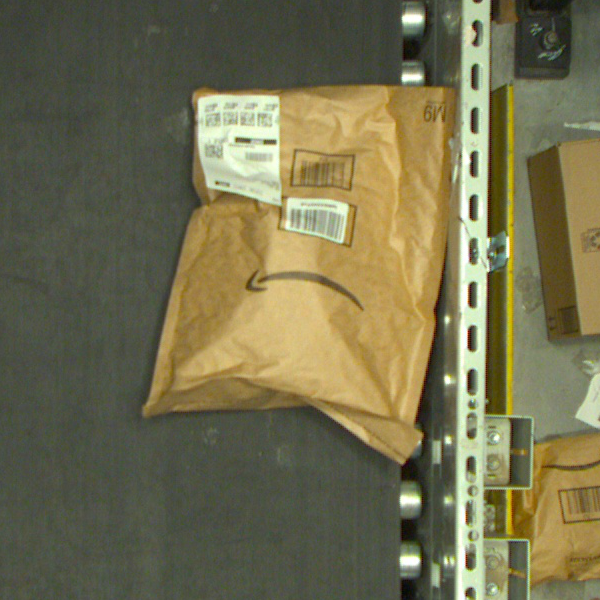}}{\footnotesize{RGB}}%
        \hfill%
        \stackunder[1pt]{\includegraphics[width=0.499\textwidth, height=2.2cm]{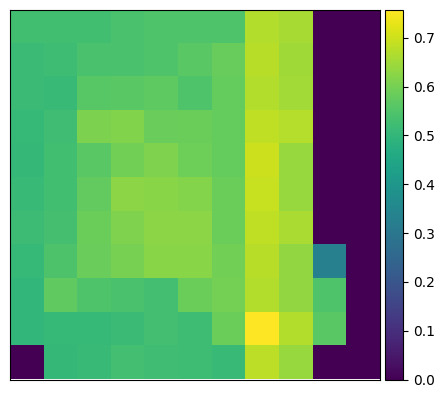}}{\footnotesize{Height Map}}%
    \end{subfigure}%
    ~~
    \begin{subfigure}{0.40\linewidth}
        \stackunder[1pt]{\includegraphics[width=0.499\textwidth, height=2.2cm]{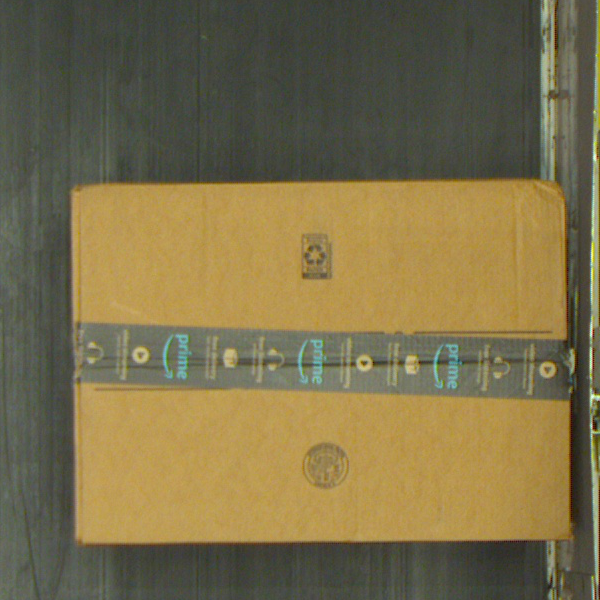}}{\footnotesize{RGB}}%
        \hfill%
        \stackunder[1pt]{\includegraphics[width=0.499\textwidth, height=2.2cm]{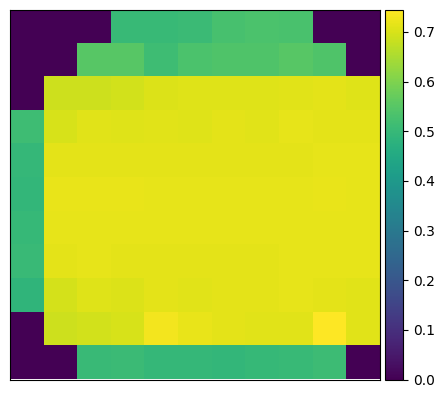}}{\footnotesize{Height Map}}%
    \end{subfigure}%
    \caption{Examples of local height maps of packages.}
    \label{fig:local-height-example}
    \vspace{-.2in}
\end{figure}

In addition to the above and other segment-specific features, we compute features that describe each segment's relationship with its surroundings, including the number of nearby segments, the \textit{adjacency graph} and the \textit{local height map} features. To compute the adjacency graph features, we construct a graph that captures the topological order of the package segments. This graph captures each detected segment's relative height with respect to its adjacent neighbor segments. Figure~\ref{fig:adj-graph} shows an example where the numbers represent the relative position ranking of the segment among its neighbors. To compute the local height map, we select point cloud within $d$ meters vicinity of the target package, and compute the height map with the coordinates of the points along the world $Z$ axis. The local height map provides detailed spatial and geometry context of the target package as well as its surrounding environment. Figure~\ref{fig:local-height-example} shows the cropped RGB image of an package and the corresponding local height map.

{\bf Training Data Generation:} The training data is generated via a 2 step process: 

\begin{enumerate}
    \item[i)] First, we apply controlled noise to an executed pick action $a_i$ to obtain a perturbed action $a_{i+1}$;
    
    \item[ii)] Then, we observe the success probability using model $F$ both before and after applying the noise: 
    \begin{equation*}
        p_i = F(\phi(s_t,a_i)), \quad p_{i+1} = F(\phi(s_t, a_{i+1})).
    \end{equation*}
\end{enumerate}

To generate the perturbed action $a_{i+1}$, we sample Gaussian noise $\varepsilon_a$ with zero mean and predetermined variances ($\sigma_{pos}$ for positions and $\sigma_{rot}$ for rotation).

For each pair consisting of a perturbed action $a_{i+1}$ and its original pick action $a_i$, we define a noise term $\delta_i$ that represents the desired gradient for optimizing pick success probability.
\begin{equation}
\delta_i = \begin{cases}
a_i - a_{i+1}, & \text{if } p_i > p_{i+1} \\
a_{i+1} - a_i, & \text{otherwise}
\end{cases}
\end{equation}
Each noise term is paired with a corresponding feature vector $\phi_i$: 
\begin{equation}
\phi_i = \begin{cases}
\phi( s_t, a_{i+1} ) & \text{if } p_i > p_{i+1} \\
\phi( s_t, a_i ) & \text{otherwise}
\end{cases}
\end{equation}

\begin{wrapfigure}{r}{0.4\textwidth}
    \centering
    \includegraphics[width=0.99\linewidth]{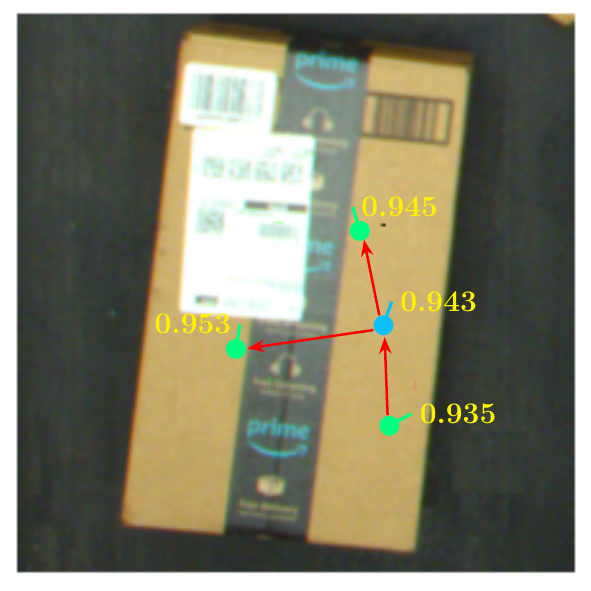}
    \vspace{-.3in}
    \caption{An illustration of the data generation approach. An initial action $a_i$ (blue) and the $N$ generated perturbed actions (green) are shown along with their predicted success probability scores. The vectors correspond to the noise term $\delta_i$ that the proposed model is learning.}
    \label{fig:data-generation}
    \vspace{-0.1in}
\end{wrapfigure}

The pairs $(\phi_i, \delta_i)$ constitute the training dataset derived from 1,000 robot picks sampled from real-world picking experiments in testing workcells. For each real-world executed pick $a_i$, we also generate $N$ perturbed actions by adding Gaussian noise. This process yields the feature and target pairs described above. The resulting dataset contains 27,977 data points split into 22,381 training data points and 5,596 testing data points. 

Figure~\ref{fig:data-generation} illustrates the described data generation approach.

\noindent {\bf Model Architecture and Training} This work trains regression models $G$ that predict action noise/gradient terms $\delta$ based on features $\phi(s_t,a)$. We employ an autoregressive approach that predicts each pick action dimension separately, while considering information from other pick action dimensions. This results in 3 distinct models corresponding to the 3 dimensions of the pick action: $X$ and $Y$ coordinates of the EoAT, and the rotation angle $r$ about the vector normal to the package surface. Note that the normal vector and $Z$-coordinate of the pick are computed based on the surface below the EoAT at the given $X$ and $Y$ coordinates.

\begin{figure}[b]
\centering
    \includegraphics[width=\linewidth]{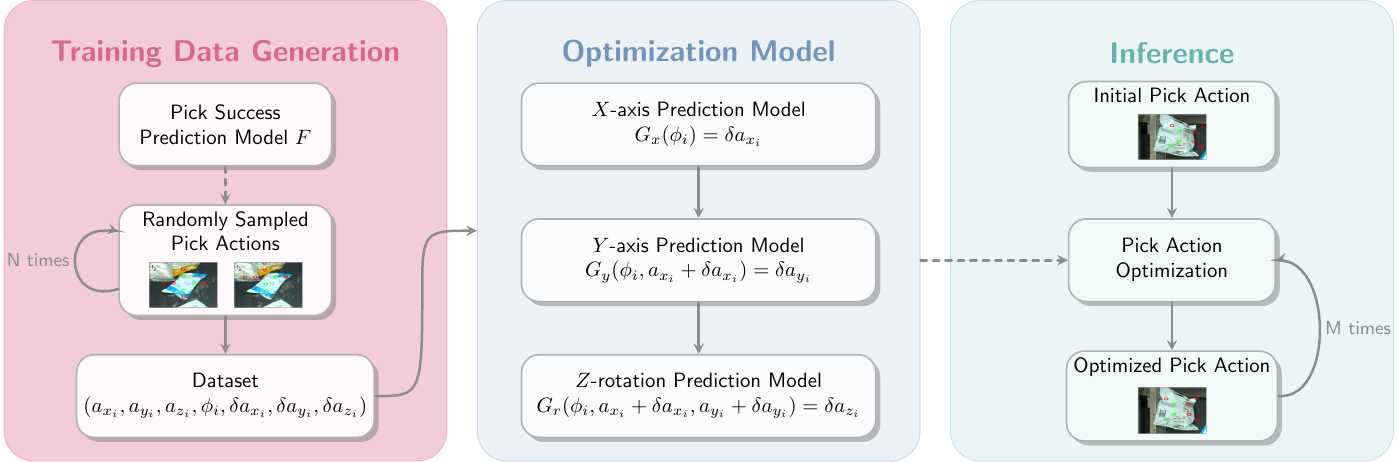}
    \caption{\small The overview of the system.}
    \label{fig:system_overview}
    \vspace{-.1in}
\end{figure}

The models are structured as follows: the $X$-axis translation model $G_x$ uses the base features $\phi_i$; the $Y$-axis translation model $G_y$ incorporates both $\phi_i$ and the predicted $X$ value; and the angular rotation model $G_r$ utilizes $\phi_i$ along with predicted $X$-axis and $Y$-axis values. This sequential prediction strategy ensures that each subsequent prediction can leverage information from previously predicted parameters, potentially capturing important spatial relationships in the picking task.

Figure~\ref{fig:system_overview} illustrates the design overview of the described system.



%% file: sections/03_experiments.tex
\section{Experiments} 
\label{sec:experiments}

We evaluated the proposed pick optimization method via using test workcells that resemble real-world logistics operations in  fulfillment centers. This section details the experimental setup, presents quantitative results, and analyzes our findings.

\begin{wrapfigure}{r}{0.37\textwidth}
\centering
  \vspace{-.2in}
    \begin{subfigure}[b]{.2\textwidth}
        \includegraphics[width=\linewidth]{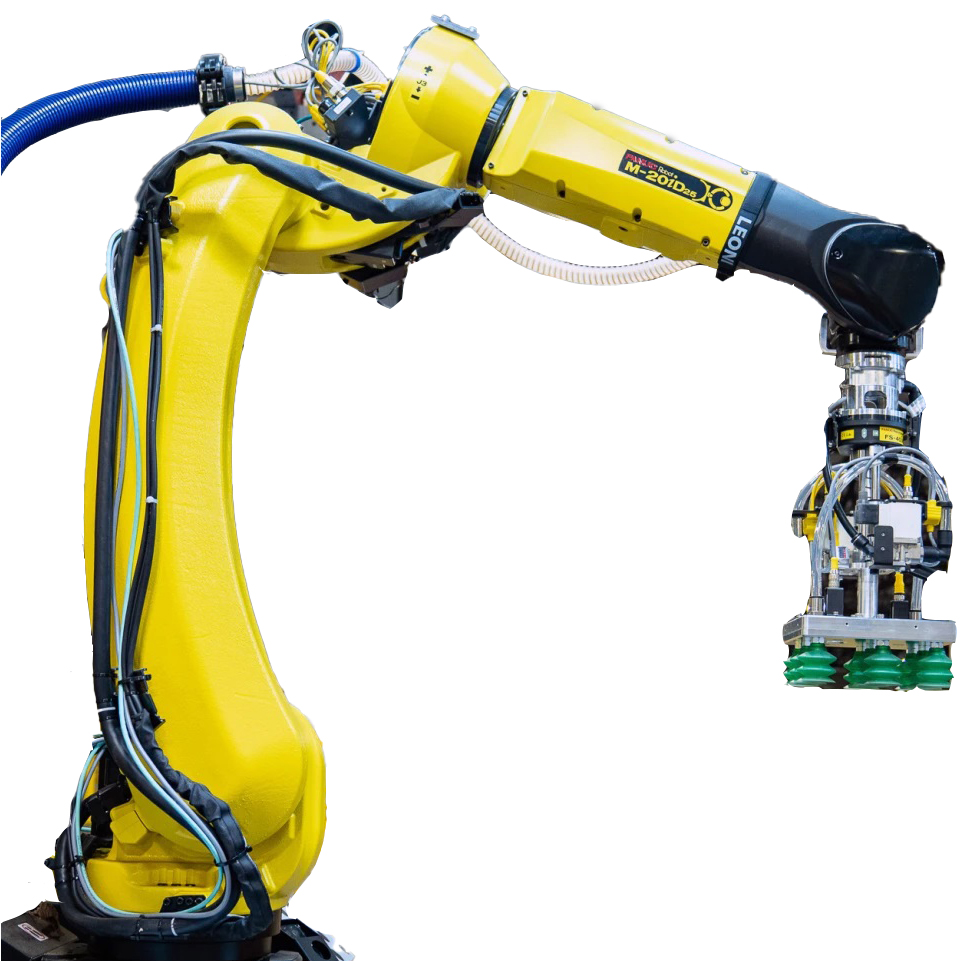}
        \caption{~}
        \label{fig:robin:robot}
    \end{subfigure}%
    ~
    \begin{subfigure}[b]{.19\textwidth}
        \centering
        \includegraphics[width=0.9\linewidth]{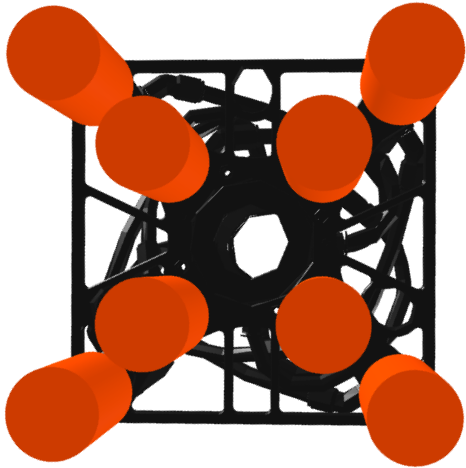}
        \caption{~}
        \label{fig:robin:eoat}
    \end{subfigure}%
\vspace{-.1in}
    \caption{\small Amazon's Robin configuration used in our experiments. (a) Arm. (b) Bottom view of the EoAT.}
\vspace{-.2in}
    \label{fig:robin}
\end{wrapfigure}

\noindent {\bf Robotic System Architecture} The test workcells resemble logistics operations, such as Amazon's Robin robotic system~\cite{robin}, which has been developed for package manipulation (Fig.~\ref{fig:robin}). The test workcells make use of a FANUC M-20iD/35 industrial robotic arm with six degrees of freedom, providing maximum operational flexibility, a $35~kg$ maximum payload capacity, and a $1831~mm$ operational reach. The end-of-arm tool (EoAT) features a vacuum-based system with 8 independently controllable suction cups arranged in an "X"-pattern configuration, covering a $25~cm \times 25~cm$ effective pick area.

\begin{wrapfigure}{r}{0.5\textwidth}
\vspace{-.25in}
\centering
\small
\renewcommand{\arraystretch}{1.2}
\begin{tabular}{|c|c|c|}
\hline
\rowcolor[HTML]{C0C0C0} 
\textbf{Pick Parameter} & \textbf{Grad. Boosting} & \textbf{2-Layer MLP} \\
\hline
\cellcolor[HTML]{EFEFEF}$p_x$ (meters) & 0.0248 & 0.0277 \\
\hline
\cellcolor[HTML]{EFEFEF}$p_y$ (meters) & 0.0254 & 0.0319 \\
\hline
\cellcolor[HTML]{EFEFEF}$r_z$ (radians) & 0.2971 & 0.3293 \\
\hline
\end{tabular}
\captionof{table}{RMSE Comparison Across Models}
\label{tab:rmse_comparison}
\vspace{-.25in}
\end{wrapfigure}


\noindent {\bf Model Architecture} Our study evaluates and compares two distinct model architectures: a gradient boosting algorithm and a 2-layer Multi-Layer Perceptron (MLP). The MLP architecture consists of two hidden layers with 35 and 2 neurons respectively, and utilizes default parameters, including the Adam optimizer with a learning rate of 0.001. We assess both models using the Root Mean Square Error (RMSE) metric on the testing dataset, calculating the RMSE for each of the three pick parameter dimensions. The comparative performance results of both models are presented in Table.~\ref{tab:rmse_comparison}.

\noindent {\bf Quantitative Results} To evaluate the effectiveness of our learned pick optimization approach, we conducted comprehensive A/B testing experiments comparing two distinct groups. {\it Control group (C):} Baseline performance using our nominal heuristic-based pick planning algorithm with a learned pick ranker (similar to Li et al.~\cite{Li:2023:rss:pickranking}). {\it Treatment group (T):} Enhanced performance using our optimized models and algorithms.


\begin{table}[!htb]
\vspace{-.15in}
\centering
\small
\renewcommand{\arraystretch}{1.2}
\begin{tabular}{|c|c|c|c|}
\hline
\rowcolor[HTML]{C0C0C0} 
\textbf{Group} & \textbf{Missed Picks} & \textbf{Missed Picks Mean Rate} & \textbf{Missed Picks 95\% CI} \\
\hline
\cellcolor[HTML]{EFEFEF}C & 22,310 & 2.23\% & [2.23\%, 2.23\%] \\
\hline
\cellcolor[HTML]{EFEFEF}T & 18,015 & 1.80\% & [1.79\%, 1.81\%] \\
\hline
\end{tabular}
\vspace{0.1in}
\caption{Comparison across Treatment Groups (1M inducts per treatment group)}
\label{tab:ab_exp_comparison}
\vspace{-0.35in}
\end{table}

One million picks from testing workcells were used for evaluation purposes for each of the treatment groups, i.e., 1~million picks for group C and 1~million picks for group C. The primary evaluation metric was missed pick rate, which measures the frequency of failed package transfers from the conveyor belt to the designated destination region. Table \ref{tab:ab_exp_comparison} provides the corresponding statistics indicating a significant, statistically significant reduction of missed pick rate by the proposed pick optimization process. 

In addition to the missed pick rate, we evaluated two additional metrics 1) infeasible pick rate, when none of the generated picks are feasible due to motion feasibility or collision, and 2) multi-pick rate, when the robot picks multiple items simultaneously. The results of these two metrics are shown in Table \ref{tab:ab_exp_comparison_additional_metrics}.

\begin{table}[!htb]
\vspace{-.15in}
\centering
\small
\renewcommand{\arraystretch}{1.2}
\begin{tabular}{|c|c|c|c|c|}
\hline
\rowcolor[HTML]{C0C0C0} 
\textbf{Group} & \makecell{\textbf{Infeasible Pick}\\\textbf{Mean Rate}} & \makecell{\textbf{Infeasible Pick}\\\textbf{95\% CI}} & \textbf{Multi-pick Mean Rate} & \textbf{Multi-pick 95\% CI} \\
\hline
\cellcolor[HTML]{EFEFEF}C & 4.49\% & [4.49\%, 4.50\%] & 0.84\% & [0.84\%, 0.85\%]\\
\hline
\cellcolor[HTML]{EFEFEF}T & 4.48\% & [4.46\%, 4.49\%] & 0.89\% & [0.89\%, 0.90\%]\\
\hline
\end{tabular}
\vspace{0.1in}
\caption{Comparison across Treatment Groups (1M inducts per treatment group)}
\label{tab:ab_exp_comparison_additional_metrics}
\vspace{-0.5in}
\end{table}

%% file: sections/04_insights.tex
\section{Insights} 
\label{sec:insights}



{\bf Experimental Scale:}  The evaluation involved a comprehensive A/B experiment by utilizing 2 million physical picks from testing workcells that resemble typical setups that are operational across the logistics industry, such as Amazon's Robin system. To the best of the authors' knowledge, the performed experiment represents one of the {\bf \emph{largest-scale evaluations ever reported on physical robotic manipulation hardware}}.

\begin{figure}[!htb]
\centering
    \begin{subfigure}[b]{.24\linewidth}
        \caption*{Probability: 0.30}
        \includegraphics[width=\linewidth]{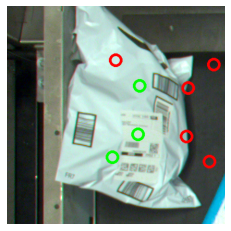}
        \label{fig:qualitative:ex1:initial}
    \end{subfigure}%
    \begin{subfigure}[b]{.24\linewidth}
        \centering
        \caption*{Probability: 0.77}
        \includegraphics[width=\linewidth]{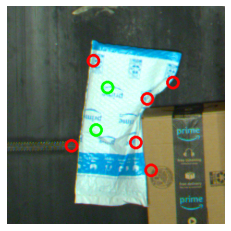}
        \label{fig:qualitative:ex1:iter-2}
    \end{subfigure}%
    \begin{subfigure}[b]{.24\linewidth}
        \centering
        \caption*{Probability: 0.95}
        \includegraphics[width=\linewidth]{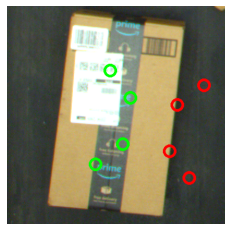}
        \label{fig:qualitative:ex1:iter-3}
    \end{subfigure}%
    \begin{subfigure}[b]{.24\linewidth}
        \centering
        \caption*{Probability: 0.94}
        \includegraphics[width=\linewidth]{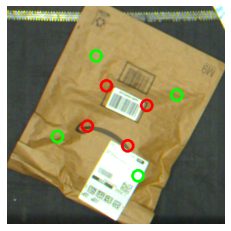}
        \label{fig:qualitative:ex1:iter-4}
    \end{subfigure}%
    
    \vspace{-.2in}
    \begin{subfigure}[b]{.24\linewidth}
        \caption*{Probability: 0.89}
        \includegraphics[width=\linewidth]{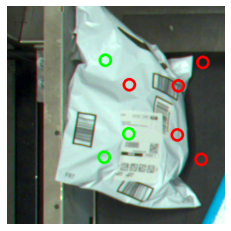}
        \label{fig:qualitative:ex2:initial}
    \end{subfigure}%
    \begin{subfigure}[b]{.24\linewidth}
        \centering
        \caption*{Probability: 0.92}
        \includegraphics[width=\linewidth]{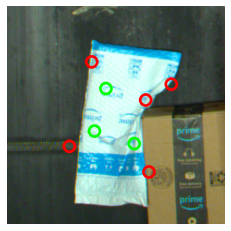}
        \label{fig:qualitative:ex2:iter-2}
    \end{subfigure}%
    \begin{subfigure}[b]{.24\linewidth}
        \centering
        \caption*{Probability: 0.97}
        \includegraphics[width=\linewidth]{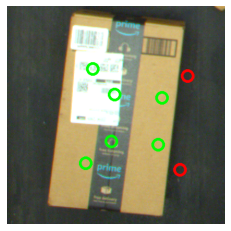}
        \label{fig:qualitative:ex2:iter-3}
    \end{subfigure}%
    \begin{subfigure}[b]{.24\linewidth}
        \centering
        \caption*{Probability: 0.96}
        \includegraphics[width=\linewidth]{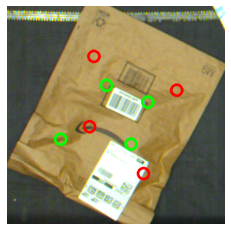}
        \label{fig:qualitative:ex2:iter-4}
    \end{subfigure}%

    \vspace{-.2in}
    \begin{subfigure}[b]{.24\linewidth}
        \caption*{Probability: 0.90}
        \includegraphics[width=\linewidth]{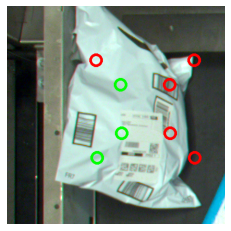}
        \label{fig:qualitative:ex3:initial}
    \end{subfigure}%
    \begin{subfigure}[b]{.24\linewidth}
        \centering
        \caption*{Probability: 0.92}
        \includegraphics[width=\linewidth]{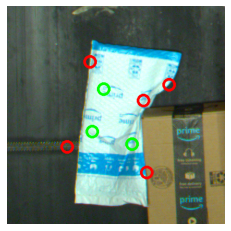}
        \label{fig:qualitative:ex3:iter-2}
    \end{subfigure}%
    \begin{subfigure}[b]{.24\linewidth}
        \centering
        \caption*{Probability: 0.98}
        \includegraphics[width=\linewidth]{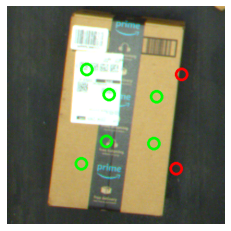}
        \label{fig:qualitative:ex3:iter-3}
    \end{subfigure}%
    \begin{subfigure}[b]{.24\linewidth}
        \centering
        \caption*{Probability: 0.95}
        \includegraphics[width=\linewidth]{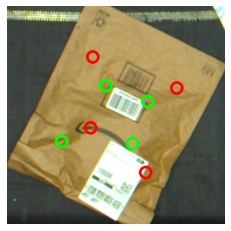}
        \label{fig:qualitative:ex3:iter-4}
    \end{subfigure}%
    \vspace{-.15in}
    \caption{\small Qualitative examples of pick optimization. The circles show the location of suction cups projected over the scene. Red is for retracted, deactivated suction cups. Green corresponds to extended, activated cups. The probabilities on top of each image correspond to the reported probabilities estimated of the Pick Success Prediction (PSP) model $F$. (first row) Initially sampled pick action. (second row) Optimized action after 2 iterations of the proposed approach. (third row) Optimized action after 3 iterations of the proposed approach.}
    \label{fig:inference-qualitative-example}
    \vspace{-.25in}
\end{figure}

{\bf Qualitative Results:} Figure~\ref{fig:inference-qualitative-example} provides qualitative examples from the physical evaluation of the proposed machine learning approach for optimizing robotic pick parameters in manipulation tasks. The method draws inspiration from flow matching techniques \cite{lipman2023flow}, where machine learning models learn to map initial pick parameters to their optimal values through a learned vector field. The proposed strategy learns such a vector field via supervision given a previously trained Pick Success Prediction (PSP) model $F$. The proposed framework could be adapted to accommodate alternative objectives, such as incorporating human preferences, through modifications to the training data generation process, such as leveraging human annotations and scores on the picks. 

The examples above highlight that training a vector field in this manner allows the proposed method to iteratively refine the initial sampled pick actions (left images) to progressively improve their predicted success probabilities (after 2 and 3 iterations). On the top row, the picks always activate 3 suction cups but the position of the suction cups is incrementally adapted by the model so that eventually the cups are pressed against surfaces of the package with improved attachment - and away from the label, which may be ripped of by the suction cup during transfer. On the bottom row, the initial sampled pick had only 2 suction cups activated but the model is able to optimize the pick to allow for 3 active suction cups, which typically means a more robust pick. 


{\bf Statistical Significance and Impact on Operations:} The evaluation focuses on analyzing missed pick events, which occur when packages fail to be successfully placed on mobile robots, as detected by an integrated perception system. The A/B experiments revealed statistically significant reduction in missed pick rates compared to alternatives, underscoring the potential of pick optimization via ML in improving real-world robotic manipulation performance over time. When a missed pick occurs, one of two scenarios unfolds: (i) either the package remains or falls onto the conveyor, requiring a re-pick attempt and reducing the workcell's throughput and operational efficiency, or (ii) it results in an ``amnesty" case when it is dropped outside the conveyor belt. Amnesty cases incur even greater costs due to the necessity of human intervention as the packages may block the workcell operation and may imply package loss or damage.


Results presented in Table ~\ref{tab:ab_exp_comparison} demonstrate that group T achieved significantly lower missed pick rates compared to group C. The statistical significance of this improvement is confirmed by the non-overlapping 95\% confidence intervals between the two groups. The absolute missed pick rate reduction of 0.43 percentage points represents a substantial 19.25\% relative improvement compared to the control group (C). 
The statistically significant improvement achieved implies significant benefits if deployed in real production environments, where the scale of operations are even bigger than the considered testing environment. Additional failure modes exist in real production systems beyond missed picks. As shown in Table ~\ref{tab:ab_exp_comparison_additional_metrics}, we observed that there is no statistical significance between C and T1 in infeasible picks rate, while T1 shows slightly higher multi-pick rates. These additional metrics indicate that the pick optimization does not improve the pick feasibility, which is expected given the Pick Success Prediction (PSP) model was trained on predicting missed picks outcomes. On the other hand, we hypothesize that with improvements in the pick qualities, the robot becomes more capable of picking up items, which can lead to higher chance of multi-pick.



{\bf Criteria for Model Selection:} To deploy the approach on a large-scale robotic system, both software integration complexity and model capabilities need to be considered. There are straightforward software integration processes for the selected models, while maintaining sufficient model complexity for expressing picking success probabilities. As Table ~\ref{tab:rmse_comparison} shows, the gradient boosting model achieves superior RMSE performance across all 3 pick parameter dimensions compared to the 2-layer MLP, demonstrating higher accuracy in predicting optimal pick parameters. These findings led to the selection of the gradient boosting model for the evaluation of the proposed treatment against the control group. Its superior performance can be attributed to its ability to effectively handle categorical features. Its robust gradient boosting algorithm can capture complex relationships between input features and target pick parameters. The gradient boosting model's performance establishes a strong baseline against which future, more sophisticated models -potentially employing deep neural network architectures - can be compared.


{\bf Directions Forward:} While the current implementation has shown promising results, there are several opportunities for further enhancement. One limitation is the absence of rich visual information in the feature vectors. To address this, future work can explore machine learning architectures that efficiently process visual inputs, particularly through the use of image embeddings. This can enable the system to capture and utilize more comprehensive information about the manipulation environment. Furthermore, as the method is actively generating effective picks on the robot fleet, it provides valuable data on the physical world interactions. This data will be instrumental in training and evaluating more sophisticated pick success prediction models, which can subsequently further improve the system's performance resulting in a lifelong improvement cycle.



%% file: arxiv_main.bbl
\begin{thebibliography}{10}
\providecommand{\url}[1]{{#1}}
\providecommand{\urlprefix}{URL }
\expandafter\ifx\csname urlstyle\endcsname\relax
  \providecommand{\doi}[1]{DOI~\discretionary{}{}{}#1}\else
  \providecommand{\doi}{DOI~\discretionary{}{}{}\begingroup \urlstyle{rm}\Url}\fi

\bibitem{robin}
{Amazon Robotics: Robin} (2022).
\newblock \urlprefix\url{https://www.amazon.science/latest-news/robin-deals-with-a-world-where-things-are-changing-all-around-it}

\bibitem{grasp-survey}
Bohg, J., Morales, A., Asfour, T., Kragic, D.: Data-driven grasp synthesis—a survey.
\newblock IEEE TRO \textbf{30}(2), 289--309 (2014)

\bibitem{suctionnet}
Cao, H., Fang, H.S., Liu, W., Lu, C.: {SuctionNet-1Billion}: A large-scale benchmark for suction grasping.
\newblock IEEE RAL \textbf{6}(4), 8718--8725 (2021)

\bibitem{apc_paper}
Correll, N., Bekris, K.E., Berenson, D., Brock, O., Causo, A., Hauser, K., Okada, K., Rodriquez, A., Romano, J.M., Wurman, P.R.: Analysis and observations from the first amazon picking challenge.
\newblock IEEE TASE \textbf{15}, 172--188 (2018)

\bibitem{fang2023anygrasp}
Fang, H.S., Wang, C., Fang, H., Gou, M., Liu, J., Yan, H., Liu, W., Xie, Y., Lu, C.: {AnyGrasp}: Robust and efficient grasp perception in spatial and temporal domains.
\newblock IEEE TRO \textbf{39}(5), 3929--3945 (2023)

\bibitem{fang2020graspnet}
Fang, H.S., Wang, C., Gou, M., Lu, C.: {GraspNet-1Billion}: A large-scale benchmark for general object grasping.
\newblock In: CVPR, pp. 11,441--11,450 (2020)

\bibitem{dl_robot_grasping}
Lenz, I., Lee, H., Saxena, A.: Deep learning for detecting robotic grasps.
\newblock In: Robotics: Science and Systems (RSS), pp. 1--8. Berlin, Germany (2013)

\bibitem{Li:2023:rss:pickranking}
Li, S., Keipour, A., Jamieson, K., Hudson, N., Swan, C., Bekris, K.: Demonstrating large-scale package manipulation via learned metrics of pick success.
\newblock In: Robotics: Science and Systems (RSS) (2023)

\bibitem{Li:2023:iros-workshop:pick-strategies}
Li, S., Keipour, A., Jamieson, K., Hudson, N., Zhao, S., Swan, C., Bekris, K.: Pick planning strategies for large-scale package manipulation.
\newblock In: Learning Meets Model-based Methods for Manipulation and Grasping Workshop, 2023 IEEE/RSJ International Conference on Intelligent Robots and Systems (IROS), pp. 1--4. Detroit, MI, USA (2023)

\bibitem{lipman2023flow}
Lipman, Y., Chen, R.T.Q., Ben-Hamu, H., Nickel, M., Le, M.: Flow matching for generative modeling.
\newblock In: ICLR (2023)

\bibitem{affordance}
Liu, H., Deng, Y., Guo, D., Fang, B., Sun, F., et~al.: An interactive perception method for warehouse automation in smart cities.
\newblock IEEE TII \textbf{17}(2), 830--838 (2021)

\bibitem{mahler2018dex}
Mahler, J., Matl, M., Liu, X., Li, A., Gealy, D., Goldberg, K.: {Dex-Net} 3.0: Computing robust vacuum suction grasp targets in point clouds using a new analytic model and deep learning.
\newblock In: ICRA, pp. 5620--5627. Brisbane, Australia (2018)

\bibitem{graspit}
Miller, A., Allen, P.K.: Graspit! a versatile simulator for robotic grasping.
\newblock IEEE RAM \textbf{11}(4), 110--122 (2004)

\bibitem{geometric}
Morales, A., Chinellato, E., Fagg, A., del Pobil, A.: Experimental prediction of the performance of grasp tasks from visual features.
\newblock In: IROS, pp. 3423--3428 (2003)

\bibitem{gg-cnn}
Morrison, D., Corke, P., Leitner, J.: Learning robust, real-time, reactive robotic grasping.
\newblock IJRR \textbf{39}(2-3), 183--201 (2020)

\bibitem{Che:2025:rss:deeppick}
Wang, C., Vanbaar, J., Mitash, C., Li, S., Randle, D., Wang, W., Sontakke, S., Bekris, K., Katyal, K.: Demonstrating multi-suction item picking at scale via multi-modal learning pick success.
\newblock In: Robotics:Science \& Systems (RSS) (2025)

\bibitem{yuan2023m2t2}
Yuan, W., Murali, A., Mousavian, A., Fox, D.: {M2T2}: Multi-task masked transformer for object-centric pick and place.
\newblock In: CoRL, pp. 1--12. Atlanta, GA (2023)

\end{thebibliography}
